# LLM_annotate: A Python package for annotating and analyzing fiction characters

Hannes Rosenbusch (h.rosenbusch@uva.nl; 0000-0002-4983-3615)
Department of Psychological Methods, University of Amsterdam

LLM_annotate is a Python package for analyzing the personality of fiction characters with large language models. It standardizes workflows for annotating character behaviors in full texts (e.g., books/movie scripts), inferring character traits, and validating annotation/inference quality via a human-in-the-loop GUI. The package includes functions for text chunking, LLM-based annotation, character name disambiguation, quality scoring, and computation of character-level statistics and embeddings. Researchers can use any LLM—commercial, open-source, or custom—for the package's functionality. Through tutorial examples using The Simpsons Movie and the novel Pride and Prejudice, I demonstrate the usage of the package for efficient and reproducible character analyses.

## LLM_annotate: A Python package for annotating and analyzing fiction characters

The impact of a story depends on its characters (Bourrier & Thelwall, 2020; Cohen, 2013; Raney, 2004; Šauperl, 2012; Tchernev, 2022). Readers might relate to their book's protagonist, hate the antagonist, or fall in love with a quirky sidekick. Similarly, TV audiences might laugh, cry, emulate, or refuse to spend any more time with the characters on the screen. Accordingly, research from behavioral, aesthetic, and computational disciplines often analyzes character behaviors and attributes (e.g., Forster, 1985; Gottschall et al., 2003; Priyatno et al., 2024; Yuan et al., 2024; Yu et al., 2023). Here, I introduce a small Python package, LLM_annotate, that standardizes the use of LLMs for the purpose of annotating character behaviors and scoring character traits. The package consists of functions to deliver raw annotations, disambiguate character names, collect character-level statistics, and provide a GUI for human-in-the-loop validation. After a brief review of character annotation methods, I will demonstrate the use of the package, point out best practices, and discuss shortcomings and possible extensions.

### Annotating character behaviors and traits

Researchers annotate behaviors and traits of fictional characters to analyze historical developments (Underwood et al., 2018), predict authorship (e.g., Wolyn & Simske, 2023), review societal norms (e.g., Abbasi et al., 2023), investigate the psychology of authors and readers (e.g., Kokesh & Sternadori, 2015; Mishara, 2010), and map reader preferences to book contents (e.g., Johnson et al., 1984; Krakowiak, 2015). The practice of analyzing character attributes has become so central in literary studies that researchers have started building character databases for books and movie scripts (Sang et al., 2022; Yu et al., 2023).

Commonly, researchers are interested in a set of attributes or behaviors among characters in a book or other media piece, and thus have human annotators generate the desired annotations (Douglas, 2008; Faber & Mayer, 2009; Hesse et al., 2005). Given the high cost of such annotations, computational researchers have developed a range of methods to (partially) automate this process. For instance, past work relied on predefined word lists to analyze character attributes (e.g., Tausczik & Pennebaker, 2010; Mohammad & Turney, 2013). Flekova & Gurevych (2015) used such lists to predict personality traits of fictional characters with above chance accuracy. Other researchers have used extensions of this method by factoring in context for the word occurrences (e.g., is a relevant verb performed by the character or towards the character; Bamman et al., 2013; 2014).

While word-level features and count-based analyses have been of great use to literary researchers, they are limited in the types of annotations they afford. Some character behaviors and attributes cannot be (reliably) inferred with such methods because the indicators are too varied, context-dependent, and intertwined with indicators of related attributes. Take the sentence "Dave served himself before handing the bowl to his grandmother; his father shook his head." While the text suggests that the father is authoritative and mindful of table manners, it is difficult to measure these specific attributes with a word list (cf. Jaipersaud et al., 2024). Thus, researchers working on authoritarianism or table manners would likely have to resort to expensive human-generated annotations.

As an alternative, researchers have trained custom machine learning models, most notably deep neural networks, to annotate character attributes based on full text inputs (as opposed to word counts). For instance, Martín Sujo and colleagues finetuned a large neural network to annotate Myers-Briggs personality types of characters in short stories (2022). However, many researchers lack the time, expertise, and/or labeled data to train a custom neural network for their annotation needs.

Fortunately, the most prominent group of neural networks–Large Language Models (LLMs)–are usually multi-purpose models, meaning that they can carry out text-based tasks without being specifically trained to, for instance, annotating authoritarian behavior of fictional characters. Jaipersaud and colleagues compared the accuracy of LLM-based character annotations with specialized neural networks and found the former to be superior in annotating characters' intelligence, appearance, and power (2024). Similarly, Yuan and colleagues demonstrated promising performances of LLMs in annotating attributes, personality, and relationships of characters (2024).

**LLM_annotate**

While lightning-fast LLM annotations are tempting for anyone who has gone through the process of manually collecting and labeling character behaviors, an automatic LLM pipeline needs to include quality assurances. In the following, we introduce common problems with (LLM) annotations and how they are addressed in the LLM_annotate package.

Annotators (human or LLM) may produce inaccurate labels because their instructions were insufficient or lacked clarity. It is relatively simple to annotate a character's height or hair color, but annotating their 'openness to experience' is fairly challenging, especially if the concept is not or insufficiently explained, or if annotators disagree about behavioral indicators. To induce deliberate instructions and explicitness, the LLM_annotate package requires users to define the traits that they want annotated, and provide examples of positive and negative indicators (arguments to function 2 in Figure 1).

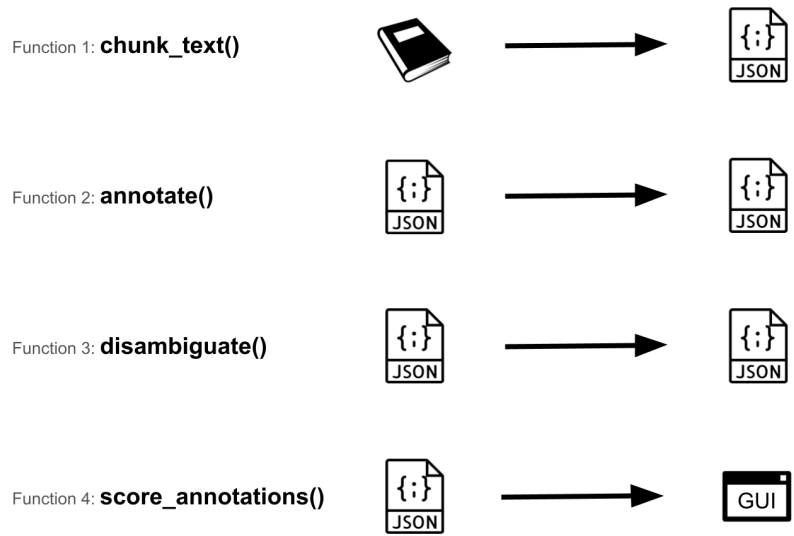

**Figure 1.** Four key functions of the LLM_annotate pipeline. Additionally, the package includes a function to compute character-wise statistics (e.g., mean trait scores, credible intervals), and another function to generate numerical, high-dimensional character representations (through the use of an embedding model).

Further, LLMs (and human annotators) may misattribute character actions and attributes. For some works of fiction—particularly when characters change names, disguise themselves, or when authors employ highly complex or ambiguous syntax or plot structures—it can be difficult to determine who performed which action. LLM_annotate addresses this challenge in two ways. First, it automatically reviews the full list of characters for pseudonyms across varying section lengths and with cross-section memory (see third function in Figure 1). Second, the human expert is prompted to review and, if necessary, correct the names that the model attempts to merge under the same character.

Additionally, LLMs might misinterpret behaviors and thus label them with the wrong traits (e.g., a big brother pretending to lose a game of chess indicates benevolence rather than lack of skill). Such cases may arise from particularly challenging inputs or LLM hallucinations (for a review, see Huang et al., 2025). The primary safeguard against such erroneous annotations is to quantify annotation accuracy via human cross-referencing. Quantifying interrater agreement is already a common practice for human-generated annotations, and it is increasingly important in LLM_annotate as some LLMs deviate markedly from human annotation patterns (Yuan et al., 2024). Accordingly, LLM_annotate includes a GUI for scoring the quality of LLM outputs and quantifying LLM-human agreement (see function four in Figure 1).

The following tutorial sections provide a step-by-step guide to using the LLM_annotate package. All code, data, and results can be found on its GitHub page: https://github.com/hannesrosenbusch/LLM_annotate

**Basic functions, minimal pipeline, and default parameters**

I will demonstrate the basic usage of the LLM annotate package by annotating the behaviors and traits of the characters in the Simpsons Movie (script by Moviepedia, n.d.). Using LLM_annotate's four core functions (see Figure 1) with default parameters mirrors a scenario where a researcher wants to *explore* character personalities without limiting annotations to specific traits or characters.

**Step 1: Chunking the text**

First, the full work (here: movie script) should be segmented into individual chunks like so:

```python
with open('script.txt', 'r', encoding='utf-8') as f:
    script = f.read()

chunk_text(script, outputfile='chunks.json')
```

The reasons for splitting the text into smaller sections are as follows: Most current LLMs have a maximum input size that will fall short of a complete book or movie script. As researchers can use LLM_annotate with any LLM, they can adjust the context size to avoid processing errors. Further, we observed that more character behaviors are collected, if the annotation occurs a few pages at a time (although it heightens the need for name disambiguation in Step 3). The chunk length is adjustable and set to 500 tokens by default (as measured by the tiktoken encoding "cl100k_base"). The function locates the last full sentence within the target length and splits the text there to minimize information loss. The last three sentences of the previous section are prepended to every section (except the first section) to ease the annotation of ambiguous events. The generated json file looks like this (snippet; two abbreviated chunks):

```json
{
 "1": "\nThe Simpsons Movie\n[Ralph sings Fox signature and the pictures panoramas to the moon where a spaceship lands and Scratchy comes out.]\nScratchy: We come in peace for cats and mice everywhere [...]",

 "2": "Homer: Well, I hate going. Why can't I worship the Lord in my own way, by praying like hell on my deathbed? [...]"
}
```

Naturally, researchers can also skip the use of the chunking function if they prefer to segment the text themselves or set a custom splitting token (see section below: 'Specifying characters, traits, and other function arguments').

**Step 2: Annotation**

Collecting character behaviors and labeling them with inferred traits constitutes the main step within LLM_annotate and can be accomplished like so:

```
annotate(chunkfile='chunks.json', outputfile='annotations.json')
```

When used without any optional arguments, the annotate function simply loops through the text chunks and lists actions, statements, thoughts, and prominent omissions of characters and adds a list of inferred traits to each entry. By default, OpenAI's GPT-4o model is used (given its promising performance for character annotation; Jaipersaud et al., 2024; Yuan et al., 2024) but can be exchanged for any other LLM (see below). Annotations are combined character-wise and written to a json file that looks like so:

```
{
 "Itchy": [

   {

     "Action": "Violently attacks Scratchy with an American flag, then frames him as a threat and nukes the moon to eliminate him.",

     "sadistic": 1,

     "Chunk": 1

   },

   {

     "Action": "Violently attacks Scratchy with an American flag, then frames him as a threat and nukes the moon to eliminate him.",

     "manipulative": 1,

     "Chunk": 1

   }

 ],...

 "Homer Simpson": [

    {

     "Action": "Says he'd rather pray on his deathbed than attend church, mocks religious attendees, and later whispers 'gay' repeatedly while Ned confesses",

     "Irreverent": 1,

     "Chunk": 2

    }...

   ]

}
```

The generated annotation file can be manually edited (e.g., a researcher might prefer that Itchy's various aggressions are listed as individual actions); however, the following steps in the package are specifically designed to process and evaluate the annotations file.

**Step 3: Character disambiguation**

Despite instructions to name characters by their full name and a persistent name memory across sections, LLM_annotate might produce separate lists for names that pertain to the same character (e.g., when names change/differ across sections). This problem also occurs when using multiple human annotators and requires the merging of the respective action/inference lists. In LLM_annotate, this can be achieved like so:

```
disambiguate('annotations.json', 'refined_annotations.json', 'chunks.json')
```

The disambiguate function iteratively loops through the list of characters in the annotations file, and tests name combinations for possible pseudonyms by reviewing multiple chunks and larger sections from the full text where one or both of the characters have annotated behaviors. If actions attributed to one character name can be consistently attributed to the pseudonym, the respective lists of annotations are merged. Crucially, the function prints the merged characters for the user like so (hypothetical output):

```
Initial pseudonym lists from AI:

"Homer" "Homer Simpson"

"Wiggum" "Police chief"

"Itchy" "Scratchy"
```

A researcher likely disagrees with the suggestion that Itchy and Scratchy are the same character. In that case, rather than manually adjusting the annotations.json file, they can re-run the disambiguation function and specify the desired merges like so:

```
disambiguate('annotations.json',
    'refined_annotations.json',
    'chunks.json',
    list_of_pseudonym_lists = [["Homer" "Homer Simpson"]["Wiggum" "Police chief"]])
```

This argument skips all LLM evaluations and limits the character merging to the user-specified lists.

**Step 4: Quality assessment**

The final necessary step within LLM_annotate is to quantify the quality of the LLM annotations through human review. For this purpose, LLM annotate provides a standardized GUI for judging individual, randomly selected annotations. The GUI can be triggered like so:

```
score_annotations('refined_annotations.json', 'eval.jsonl', 'chunks.json')
```

The GUI is shown in Figure 2. Note that the possible labels (here: "Correct", "Questionable", and "Incorrect") can be set by the user.

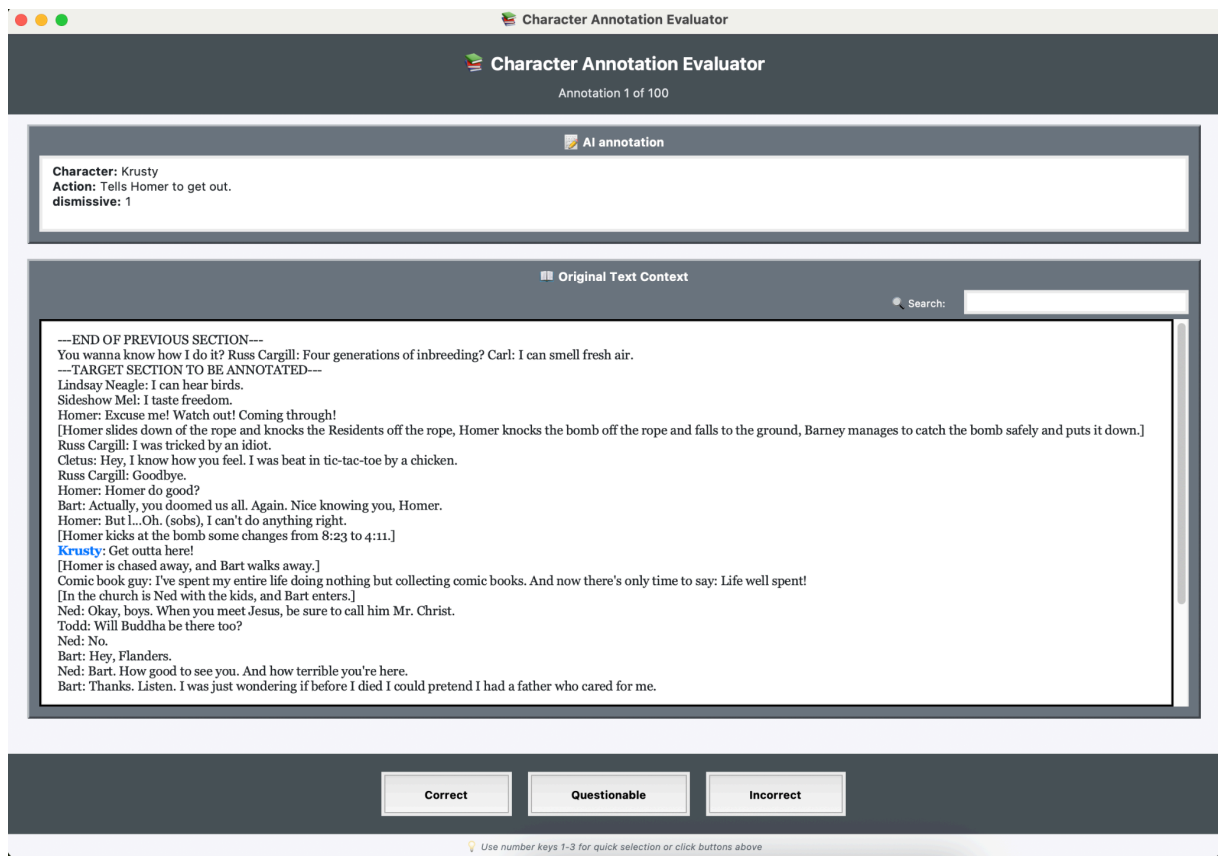

**Figure 3.** Screenshot of the human-in-the-loop GUI for evaluating LLM annotations.

After reviewing 100 annotations (the default number of the score_annotations function) alongside the respective text chunks, the user is presented with overall quality statistics as well as credible intervals. In this case, the researcher rated 95% [89%-98%] of the annotations as completely correct, 4% [2%-10%] as questionable, and 1% [0%-5%] as incorrect. Note, that these judgments cover correct action attribution and trait inference, but they do *not* score omissions (i.e., relevant actions that were missed by the annotate function. Such cases require manual review of the annotations and/or a sensitivity analysis (see below).

**Step 5: Optional analyses**

LLM_annotate includes optional functions for generating statistics, plots, and embeddings for individual characters. In case of exploring The Simpsons Movie, the researcher might, for instance, be interested to review the action counts and accumulated attributes for individual characters, which can be obtained like so:

```
compute_annotation_statistics("refined_annotations.json", "statistics.json")
```

By default, the function produces a comprehensive output file with summary statistics and credible intervals, as well as a summary plot for the most prominent characters, in this case:

**Homer Simpson (94 total):**

humorous (7), impulsive (7), self-centered (5), irresponsible (4), reckless (4), careless (4), carefree (4), determined (4), dismissive (3), easily distracted (2), childish (2), resourceful (2), optimistic (2), stubborn (2), disrespectful (1), irreverent (1), cynical (1), affectionate (1), mischievous (1), practical (1), competitive (1), tolerant (1), adventurous (1), manipulative (1), immature (1), forgetful (1), rude (1), gluttonous (1), sentimental (1), dramatic (1), panicked (1), evasive (1), nervous (1), in denial (1), fearful (1), protective (1), sarcastic (1), apologetic (1), dependent on family (1), imaginative (1), unaware (1), transformative (1), self-sacrificing (1), confused (1), seeking redemption (1), impatient (1), critical (1), discontent (1), selfish (1), resentful (1), self-reflective (1), resilient (1), clumsy (1), well-meaning (1), respectful (1), appreciative of effort (1)

**Marge Simpson (59 total):**

concerned (9), protective (5), responsible (4), supportive (3), practical (3), cautious (2), sentimental (2), assertive (2), determined (2), caring (2), conscientious (1), socially aware (1), thoughtful (1), persistent (1), curious (1), investigative (1), tired (1), long-suffering (1), pessimistic (1), observant (1), pragmatic (1), decisive (1), forgiving (1), skeptical (1), resourceful (1), quick-thinking (1), disillusioned (1), family-oriented (1), urgent (1), directive (1), affectionate (1), romantic (1), punctual (1), maternal (1), attentive (1)

**Bart Simpson (59 total):**

rebellious (7), mischievous (7), playful (5), daring (3), vulnerable (2), observant (2), cynical (2), carefree (1), childish (1), self-indulgent (1), dramatic (1), frustrated (1), nostalgic (1), secretive (1), impulsive (1), reckless (1), self-destructive (1), disrespectful (1), cautious (1), caring (1), troubled (1), critical (1), realistic (1), supportive (1), creative (1), humorous (1), resentful (1), seeking consistency (1), forgiving (1), adventurous (1), clever (1), provocative (1), discontent (1), persistent (1), blunt (1), seeking affection (1), impatient (1), restless (1)

**Lisa Simpson (49 total):**

concerned (5), environmentally conscious (5), determined (3), persistent (2), idealistic (2), thoughtful (2), intelligent (2), cautious (2), caring (2), proactive (1), responsible (1), enthusiastic (1), talkative (1), passionate (1), logical (1), emotional (1), compliant (1), expressive (1), romantic (1), observant (1), supportive (1), encouraging (1), inquisitive (1), obedient (1), aware (1), empathetic (1), moralistic (1), practical (1), strategic (1), musical (1), playful (1), compassionate (1), conscientious (1)

**Figure 2.** Annotations for characters from the Simpsons Movie. The numbers in parentheses indicate the number of behaviors that were labeled with the given trait.

**Specifying characters, traits, and other function arguments**

Above, we demonstrated LLM_annotate for its most basic functionality: fully exploratory annotation of all characters in one or more written works. However, researchers are often interested in specific traits or characters. Such targeted labeling tasks are undertaken with the same functions as above, but under specification of additional input arguments.

In the following scenario, we simulate the annotation procedure for a hypothetical group of psychology researchers who are interested in the concept of social dominance orientation – a construct from social psychology which refers to an individual's preference for hierarchy among social groups (cf., Pratto et al., 1994; Sidanius & Pratto, 2001). The researchers are interested in the interclass relations in Jane Austen's novel Pride and Prejudice and how the characters support or rebel against class differences. Additionally, the researchers would like to use their own custom LLM for the annotation task. As before, they obtain the raw text of the book and segment it (example books are included in the package repository). However, they assume that their own LLM works better with full chapters rather than the short (~2 page) chunks that the chunk_text function produces by default. Thus, they insert a custom splitting string at the end of each chapter and run their chunking code like this:

```
with open("pride and prejudice.txt", 'r', encoding='utf-8') as f:
    novel = f.read()

chunk_text(novel,
    "chunks.json",
    custom_splitter="<SPLIT_HERE>")
```

As the researchers are unsure whether this new chapter-wise splitting is *really* a good idea, they conduct a sensitivity analysis with the default splitting parameters:

```
chunk_text(novel, "chunks_default_chunking.json")
```

While the custom, chapter-wise splitting results in 61 chunks, LLM_annotate's default splitting results in 343 chunks. Next, the researchers want to annotate behaviors of characters that give insight into their social dominance orientation. Additionally, they want to score each character on their level of agreeableness, as they hypothesize that characters who want to uphold the social hierarchy (high SDO) are presented as less agreeable in Jane Austen's work. In order to score SDO and agreeableness, the researchers have to provide an additional argument to the annotate function with definitions and example inferences of the two traits:

```
traits = {
  "agreeableness": {
    "trait_explanation": "Agreeableness refers to traits that promote social harmony and cooperation.",
    "examples": [
      {"name": "John Doe",
       "action": "Volunteers at a local homeless shelter.",
       "assessment": "Volunteering is a time-intensive and communal activity",
       "rating": 3},
      {"name": "Jane Smith",
       "action": "Refuses to help a friend in need.",
       "assessment": "In this particular scene, the character seemed rather self-centered.",
       "rating": -2}
    ]
  },
  "social dominance orientation": {
    "trait_explanation": "Social dominance orientation refers to the extent to which an individual prefers hierarchical relationships between social groups.",
    "examples": [
      {"name": "John Doe",
       "action": "Supports policies that favor the wealthy.",
       "assessment": "This action suggests that the character is rather unfazed by issues of social inequality.",
       "rating": 3},
```

```
            {"name": "Jane Smith",
             "action": "Marries someone from a lower social class.",
             "assessment": "This action indicates that Jane no longer cares as much
              about rising through social hierarchies.",
             "rating": -2}
        ]
    }
}
```

Additionally, the researchers provide the full rating scale and their custom LLM. This is done by defining a text-input/text-output function and inserting it into the annotate function like so:

```python
def custom_llm(prompt):
    response = my_llm(messages=prompt, temperature=0)
    return response.text

annotate('chunks.json',
    'annotations.json',
    traits=traits,
    rating_scale=[-3, -2, -1, 0, 1, 2, 3],
    model=custom_llm,
    book_title=book)
annotate('chunks_default_chunking.json',
    'annotations_default_chunking.json',
    traits=traits,
    rating_scale=[-3, -2, -1, 0, 1, 2, 3],
    model=custom_llm,
    book_title=book)
```

If the given LLM has parameters (like temperature) it is generally advantageous to set them in a way that maximizes reproducibility. Note that the book title is an optional argument to the annotate function and serves as context for each chunk, which can be advantageous (e.g., for famous books that are likely part of the chosen LLM's training data). Additionally, the researchers could have determined which characters should be annotated (argument: target_characters) but prefer to collect annotations for all characters and analyze the 10 most prominent characters (as measured by the number of annotations).

The resulting json file is structured identically to the annotation file in the Simpsons example above. In the test run, an apparent problem occurred when running the disambiguation function for the annotations of the default-size chunks:

```python
disambiguate('annotations_default_chunking.json',
    'refined_annotations_default_chunking.json',
    'chunks_default_chunking')
```

Specifically, the function suggests to merge the annotations for 'Jane Bennet' and 'Miss Bennet'. This appears dangerous as there is more than one Miss Bennet in the book; however, a manual review of the three annotations generated for Miss Bennet reveals that all actions indeed pertained to Jane Bennet. Had one of the actions pertained to a different character (say Elizabeth Bennet), the researchers would have had to make a small manual edit to the file annotations_default_chunking.json (here: cut and paste one annotation from the Miss Bennet list to the Elizabeth Bennet list).

In the next step, the researchers want to ascertain whether their chapter-wise chunking was indeed superior to the default chunking. Thus, they score the annotations separately. Additionally, the researchers assume that inferences of agreeableness and social dominance orientation among the Pride and Prejudice characters will often be challenging and relatively subjective compared to trait inferences in the Simpsons Movie. Thus, they decide to replace the default three-point rating scale ("Correct", "Questionable", "Incorrect") with a 6-point scale that the human rater should use to indicate their own trait inference from the presented action:

```
vals = ['-3', '-2', '-1', '0', '1', '2', '3']
score_annotations('annotations.json',
    'evaluation.jsonl',
    'chunks.json',
    labels=vals)

score_annotations('annotations_default_chunking.json',
    'evaluation_default_chunking.jsonl',
    'chunks_default _chunking.json',
    labels=vals)
```

The default chunking produced 1981 annotations whereas the custom chapter-wise chunking only led to 578 annotations. The relative accuracy of the 50 annotations from each method is plotted in Figure 3.

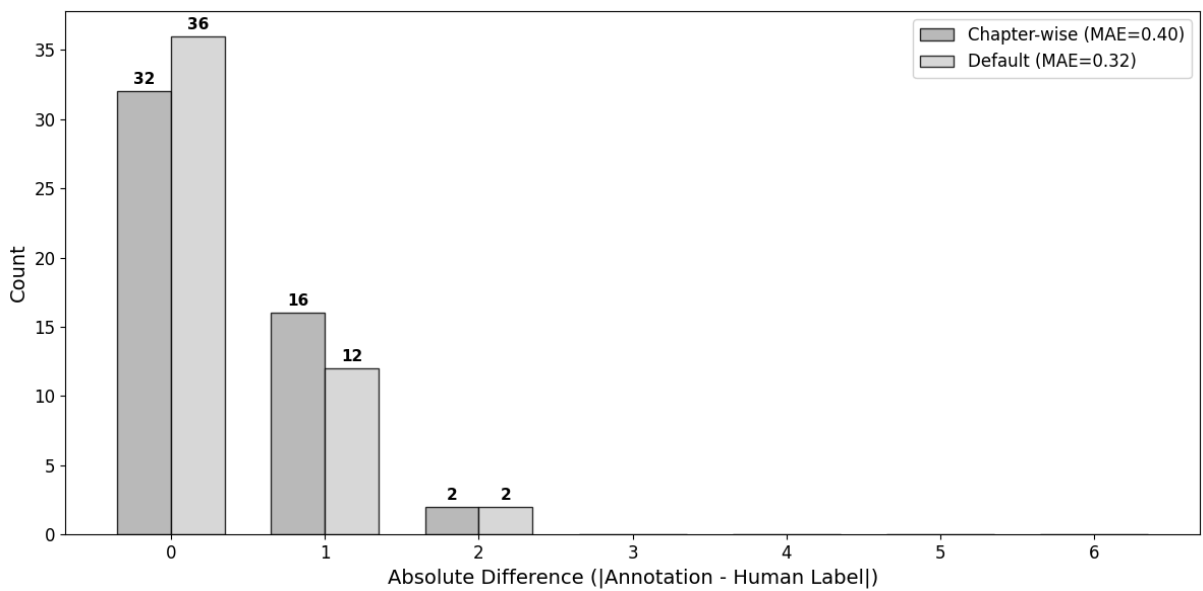

**Figure 3.** Accuracy of annotations for Pride and Prejudice characters, split by chunking method.

The accuracies entailed by the two chunking methods do not differ by much ($\chi^2(2, N = 100) = 0.81$, $p = 0.668$). Similarly, the focal analysis (i.e., the correlation between SDO and agreeableness for the ten most prominent characters) shows comparable results for both methods (default chunking: *r*(8) = -.742, *p* = .014, see Figure 4; chapter-wise chunking: (*r*(8) = -0.822, *p* = .003). Still, we advise researchers faced with decisions about parameters of LLM_annotate (here: chunk size) to conduct such sensitivity analyses, which are relatively effortless given the evaluation GUI and the highly automated workflow. Especially when deviating from a preregistration or the package's default parameters, it is worthwhile to carry parameter decisions forward to the final hypothesis tests. In this case, if the researchers *had to* decide which analysis to report, they would likely be best advised to select the default chunking method as the chapter-wise chunking produced fewer annotations and therefore less precise character-level statistics.

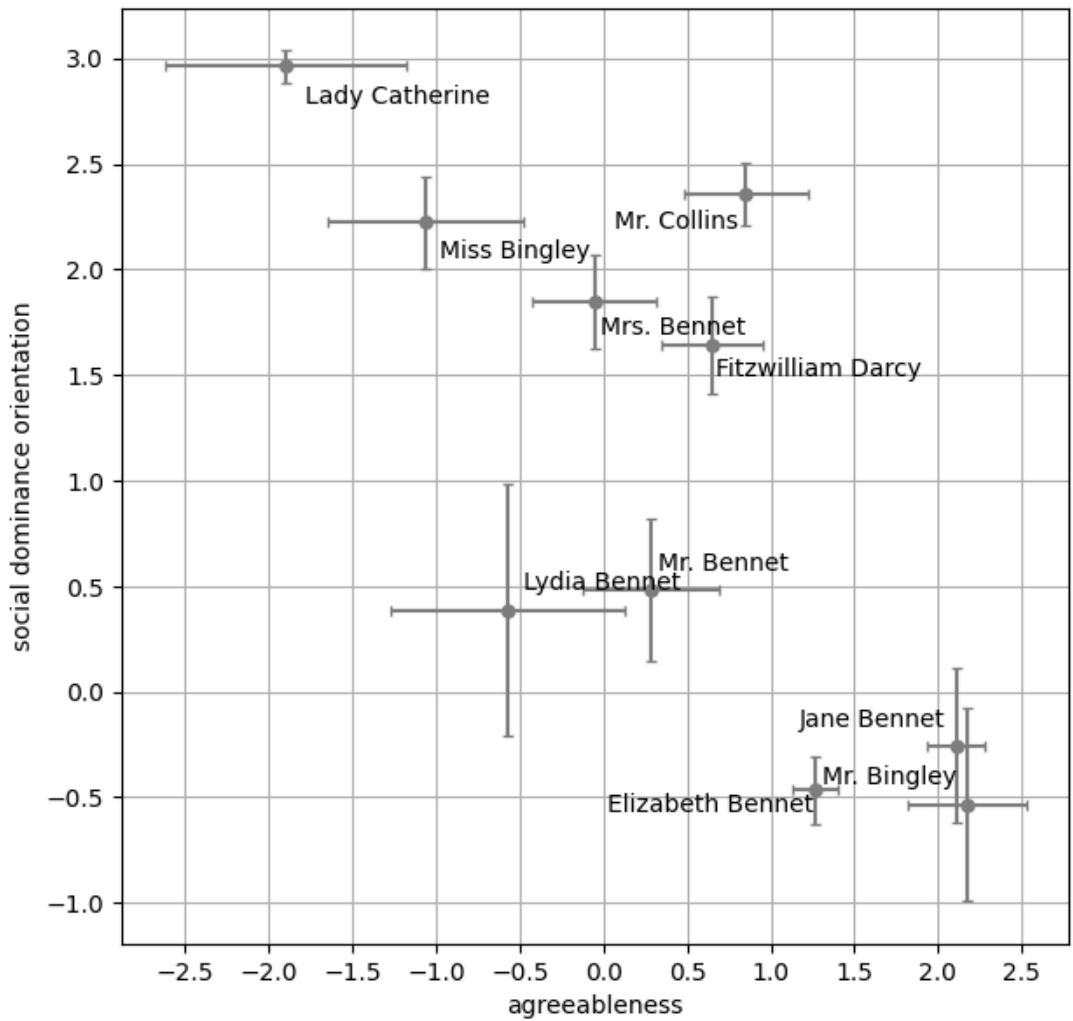

**Figure 4.** SDO and agreeableness scores of the 10 Pride and Prejudice characters with the most annotations. Whiskers indicate Bayesian credible intervals (default output of the compute_annotation_statistics function).

**Discussion**

Annotating character behaviors and inferring character traits is as essential to literary research as it is expensive. The Python package LLM_annotate provides a standardized and transparent pipeline for this task, including human-in-the-loop features and quality checks. It allows researchers to choose any large language model (LLM) they require, or even to repeat the annotation process with multiple models (e.g., for sensitivity analyses). The package is publicly available and the core functions can be extended or modified for custom usage.

Relative to human experts, LLM_annotate offers clear advantages in terms of resource efficiency and reproducibility. While previous approaches often required teams of human annotators or task-specific models (e.g., Flekova & Gurevych, 2015; Martín Sujo et al., 2022), LLM-based annotations can now be generated in a fraction of the time, with standardized outputs and transparent intermediate steps. The fully automated pipeline—from text segmentation to quality scoring—ensures that annotations can be replicated, audited, and compared across projects and models. This level of reproducibility addresses one of the major limitations of earlier annotation studies in literary research and digital humanities (Ries et al., 2024; Underwood et al., 2018; Sang et al., 2022).

However, LLM_annotate also inherits some of the known weaknesses of large language models. Long-range dependencies in narrative structure may be truncated or misinterpreted, leading to incorrect attributions of motive or personality (e.g., Severus Snape's seemingly evil behaviors in earlier Harry Potter books). As previous work has shown (Jaipersaud et al., 2024; Yuan et al., 2024), even the best-performing models may deviate from human annotation patterns, especially when traits depend on subtle contextual or cultural cues. The inclusion of expert-in-the-loop validation is therefore a necessary step when using the LLM_annotate pipeline.

In addition, researchers must still invest substantial thought before running the LLM_annotate pipeline. Conceptual definitions and examples for each trait must be carefully chosen, and appropriate scales for rating character behaviors and LLM outputs need to be determined in advance. As with any annotation procedure, the reliability of the output is contingent on the clarity of the target constructs and their indicators. The package facilitates this process by formalizing these definitions and examples in code, but interpretive rigor remains the responsibility of the researcher.

Overall, LLM_annotate contributes to a growing methodological movement toward computationally assisted interpretation of narrative characters. By combining the efficiency and scalability of LLMs with established standards of interrater validation, the package supports literary studies that are both automated and accountable.

## Acknowledgments

GPT-5 and Erdem O. Meral supported the writing of this paper. The author accepts responsibility for the correctness of the text.

LLM_ANNOTATE PACKAGE15